\documentclass[11pt,a4paper]{article}

\usepackage[hyperref]{acl}
\usepackage{times}
\usepackage{latexsym}

\usepackage{microtype}

\usepackage{amsmath}
\usepackage{amssymb}

\usepackage{caption}
\usepackage{subcaption}
\usepackage{comment}

\usepackage{pifont}% http://ctan.org/pkg/pifont

\usepackage{paralist}
\usepackage{mathtools}
\usepackage{multirow}
\usepackage[capitalise]{cleveref}
\usepackage{booktabs}
\usepackage{svg}
\usepackage[english]{babel}
\usepackage{blindtext}
\usepackage[ruled]{algorithm2e}
\usepackage{tablefootnote}
\usepackage{longtable}
\usepackage{multirow}

\newcommand{\ourtask}{NLI}

\makeatletter
\newcommand{\printfnsymbol}[1]{%
  \textsuperscript{\@fnsymbol{#1}}%
}
\makeatother

\newif\ifcomments
\newcounter{psCounter}
\newif\ifpsvar
\ifcomments
    \psvartrue
\fi
\ifpsvar
    \newcommand{\ps}[1]{{\small \color{red} \refstepcounter{psCounter}\textsf{[PS]$_{\arabic{psCounter}}$:{#1}}}}
\else
    \newcommand{\ps}[1]{}
\fi

\newcounter{ywCounter}
\newif\ifywvar
\ifcomments
    \ywvartrue
\fi
\ifpsvar
    \newcommand{\yw}[1]{{\small \color{cyan} \refstepcounter{ywCounter}\textsf{[YW]$_{\arabic{ywCounter}}$:{#1}}}}
\else
    \newcommand{\yw}[1]{}
\fi

\newcounter{pdCounter}
\newif\ifpdvar
\ifcomments
    \pdvartrue
\fi
\ifpdvar
    \newcommand{\pradeep}[1]{{\small \color{blue} \refstepcounter{pdCounter}\textsf{[PD]$_{\arabic{pdCounter}}$:{#1}}}}
\else
    \newcommand{\pradeep}[1]{}
\fi

\newcounter{mgCounter}
\newif\ifmgvar
\ifcomments
    \mgvartrue
\fi
\ifmgvar
    \newcommand{\matt}[1]{{\small \color{teal} \refstepcounter{mgCounter}\textsf{[MG]$_{\arabic{mgCounter}}$:{#1}}}}
\else
    \newcommand{\matt}[1]{}
\fi

\usepackage{glossaries}
\glsdisablehyper

\title{Generating Data to Mitigate Spurious Correlations in \\ Natural Language Inference Datasets}
\author{
  Yuxiang Wu\printfnsymbol{2}\thanks{ \hspace{5pt}  Work done while at the Allen Institute for AI.} \quad Matt Gardner \printfnsymbol{3}\textsuperscript{$*$} \quad Pontus Stenetorp \printfnsymbol{2} \quad Pradeep Dasigi \printfnsymbol{4} \\
  \printfnsymbol{2} University College London \\
    { \normalsize \tt {yuxiang.wu,p.stenetorp}@cs.ucl.ac.uk }
\\ 
  \printfnsymbol{3} Microsoft Semantic Machines
  \printfnsymbol{4} Allen Institute for AI \\
  { \normalsize \tt  mattgardner@microsoft.com, pradeepd@allenai.org}
}

\begin{document}
\maketitle

\begin{abstract}
Natural language processing models often exploit spurious correlations between task-independent features and labels in datasets to perform well only within the distributions they are trained on, while not generalising to different task distributions.
We propose to tackle this problem by generating a debiased version of a dataset, which can then be used to train a debiased, off-the-shelf model, by simply replacing its training data.
Our approach consists of 1) a method for training \textit{data generators} to generate high-quality, label-consistent data samples; and 2) a filtering mechanism for removing data points that contribute to spurious correlations, measured in terms of \textit{z-statistics}.
We generate debiased versions of the SNLI and MNLI datasets,\footnote{All our code and the generated datasets are available at \url{https://github.com/jimmycode/gen-debiased-nli}.} and we evaluate on a large suite of debiased, out-of-distribution, and adversarial test sets.
Results show that models trained on our debiased datasets generalise better than those trained on the original datasets in all settings.
% , and match the performance of a model specifically built for handling the hypothesis-only bias.
% 
On the majority of the datasets, our method outperforms or performs comparably to previous state-of-the-art debiasing strategies, and when combined with an orthogonal technique, product-of-experts, it improves further and outperforms previous best results of SNLI-hard and MNLI-hard.

\end{abstract}

\section{Introduction}

\begin{figure}
\begin{center}
\includegraphics[width=\columnwidth]{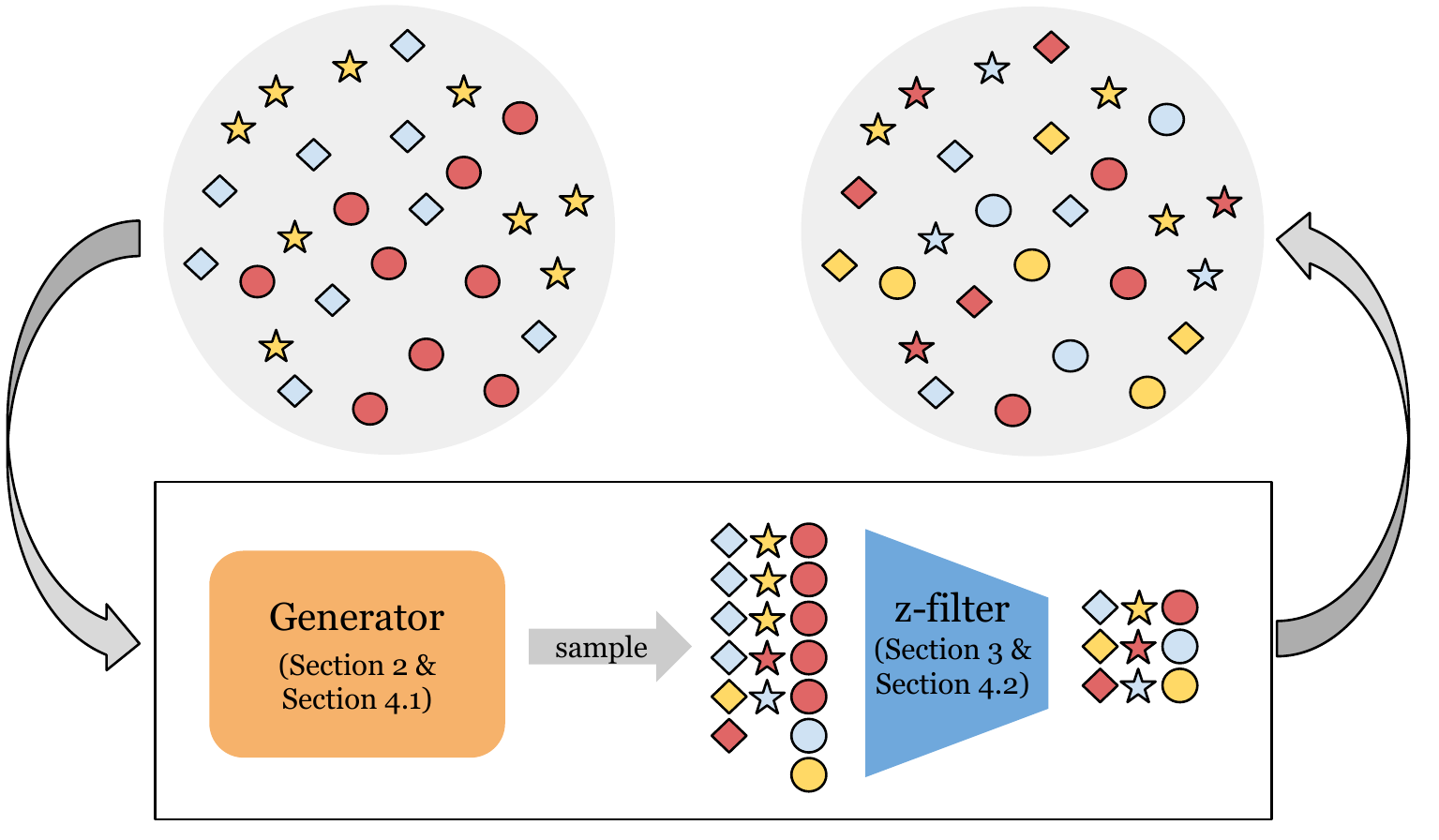}
\end{center}
\caption{Overview of our dataset bias mitigation approach. We minimise spurious correlations between labels (represented by the shapes of data points) and task-independent features (represented by their colours) with our proposed data generation pipeline.}
\label{fig:splash}
\end{figure}

Natural Language Processing~(NLP) datasets inevitably contain biases that are unrelated to the tasks they are supposed to represent.
These biases are usually artifacts of the annotation processes, task framing, or design decisions~\citep{schwartz-etal-2017-effect,geva-etal-2019-modeling,liu2021toward}. Such biases often manifest as spurious correlations between simple features of the data points and their labels~\citep{competency}. Trained models can exploit these spurious correlations to correctly predict the labels of the data points within the same distributions as those they are trained on, but fail to generalise to other distributions within the same tasks.
Consequently, the models risk modelling the datasets, but not the tasks~\citep{GururanganSLSBS18,PoliakNHRD18,hans,schuster-etal-2019-towards}.
% \ps{
%     Toned the last sentence the heck down.
%     %
%     Originally it essentially said that existing models have no generalisation power whatsoever outside of its training distribution, which is trivially false.
% }
%

We address this issue by \textit{adjusting} existing dataset distributions to mitigate the correlations between task-independent features and labels.
First, we train \textit{data generators} that generate high quality data samples in the distribution of existing datasets~(Section~\ref{sec:generator}).
Then, we identify a set of simple features that are known to be task-independent, and use the theoretical framework (i.e., z-statistics) proposed by \citet{competency} to measure correlations between those features and the labels~(Section~\ref{sec:z-score}).
Finally, we adjust the distribution of the generated samples by post-hoc filtering~(Section~\ref{sec:z-filter}) to remove the data points that contribute to high z-statistics with task-independent features, or finetuning the data generator~(Section~\ref{sec:ul-debias}) to make such data points less likely.
Unlike prior \emph{model-centric} approaches to mitigate spurious correlations~\citep{belinkov-etal-2019-dont,belinkov-etal-2019-adversarial,clark-etal-2019-dont,he-etal-2019-unlearn,karimi-mahabadi-etal-2020-end} that define new training objectives or model architectures, our approach has the advantage of keeping the objective and the model fixed, as we only alter the training data. 
% \matt{Not sure that ``advantage'' is the right term here. It's complementary. No need to give such a value judgment, and I don't really agree with the value judgment, anyway.}
%
% method

%We evaluate our method against these other yahoos on established datasets X, Y, and Z.
%We apply our method on SNLI~\citep{snli} and MNLI~\citep{mnli} datasets, which are known to contain various annotation artifacts~\citep{GururanganSLSBS18,PoliakNHRD18}. We evaluate the 

To evaluate our approach, we use the task of Natural Language Inference~(NLI), which offers a wide range of datasets~(including challenge datasets) for various domains.
We generate debiased SNLI~\citep{snli} and MNLI~\citep{mnli} distributions and evaluate the generalisability of models trained on them to out-of-distribution hard evaluation sets~\citep{GururanganSLSBS18,hans}, and the adversarial attack suite for NLI proposed by~\citet{model-agnostic-debias}.
Furthermore, we compare our method to strong debiasing strategies from the literature~\citep{belinkov-etal-2019-adversarial,stacey-etal-2020-avoiding,clark-etal-2019-dont,karimi-mahabadi-etal-2020-end,utama-etal-2020-mind,sanh2021learning,ghaddar-etal-2021-end}.
%
%For evaluation, we use the widely adopted SNLI~\citep{snli} and MNLI~\citep{mnli} datasets for both training and evaluation as they are known to contain various annotation artifacts~\citep{GururanganSLSBS18,PoliakNHRD18}.
%
%Furthermore, we measure generalisation explicitly by using the SNLI-hard~\citep{GururanganSLSBS18}, MNLI-hard~\citep{karimi-mahabadi-etal-2020-end}, HANS~\citep{hans},
%10 OOD datasets~\citep{PoliakNHRD18},
%and the adversarial attack suite for NLI proposed by~\citet{model-agnostic-debias}.

Our results show that models trained on our debiased datasets generalise better than those trained on the original datasets to evaluation sets targeting hypothesis-only biases~(by up to 2.8 percentage points) and syntactic biases~(by up to 13.3pp), and to a suite of adversarial tests sets~(by up to 4.2pp on average). 
% On the test sets targeting the hypothesis-only bias, they perform comparably to the best performing models tweaked specifically to handle the bias, but since our contributions are orthogonal to such improvements, we show that combining the two yields further improvements of up to 1.6pp.
Since our contributions are orthogonal to model-centric approaches, we show that when combined with product-of-experts~\citep{karimi-mahabadi-etal-2020-end}, our method yields further improvements and outperforms previous state-of-the-art results of SNLI-hard and MNLI-hard.
Finally, we train stronger and larger pretrained language models with our debiased datasets, and demonstrate that the performance gain by our method generalises to these larger models.

\section{Generating High-Quality Data Samples} \label{sec:generator}
% \yw{why do we need this? we need a generator to generate samples}
% 
First, we need to train a data generator $G$ to generate data samples automatically.
Our goal for the data generator is to model the true distribution as well as possible so that we can generate valid and high-quality data samples.
%Data generation is a challenging problem in itself. 
%But with the recent advances of \acrlong{plm}~\citep{gpt2,t5,bart,gpt3}, it has become a feasible task in many areas.
%

% \ps{
%     Section intro should be akin to this.
%     Why do we need this generator? To generate samples outside the original distrubiton, than preferably reflects the ``pure'' task.
% }

\subsection{Finetuning Pretrained Language Model to Generate \ourtask{} Samples}

We finetune a pretrained language model on the \ourtask{} datasets to serve as our data generator. We choose GPT-2 because it is a powerful and widely-used autoregressive language model, and it can be easily adapted to generated the premise, label, and hypothesis of an instance sequentially.
% \ps{
%     ``We could use any language model, but this one because\ldots''.
%     Too vague, try to motivate GPT-2 further.
%     Widely used, widely available, proven track record by citing others. Also, motivate a transformer-based one to begin with?
% }

Given an NLI dataset $\mathcal{D}_0$, the training objective is to minimise the following negative log-likelihood loss of generating the premise-label-hypothesis sequence, in that order:
\begin{align}\label{eq:mle}
&\mathcal{L}_{MLE} = -\sum_{i=1}^{|\mathcal{D}_0|} \log p(P^{(i)}, l^{(i)}, H^{(i)}) \nonumber \\
&= -\sum_{i=1}^{|\mathcal{D}_0|} \log p(P^{(i)}) p(l^{(i)} | P^{(i)}) p(H^{(i)}| l^{(i)}, P^{(i)}),
\end{align}
where $P^{(i)}$, $l^{(i)}$ and  $H^{(i)}$ are the premise, label and hypothesis  respectively.%
% 
% The probability is factorized in the order of premise-label-hypothesis.
\footnote{In our preliminary study, we found the factorization order premise-label-hypothesis in \cref{eq:mle} performs better than hypothesis-label-premise and premise-hypothesis-label.}

\subsection{Improving Data Generation Quality} \label{sec:data-quality}

We find that samples generated by a generator trained with only $\mathcal{L}_{MLE}$ often contain ungrammatical text or incorrect label. In this section, we introduce two techniques to improve data quality.

% \ps{Yes, just very briefly re-iterate what we stated at the start of the mani section about it needing to be related to the ``pure'' task, etc.}

\subsubsection{Unlikelihood Training to Improve Label Consistency} \label{sec:genv1_ul}

% We find that a generator trained with only $\mathcal{L}_{MLE}$ has poor \emph{label consistency}.
We observe poor \emph{label consistency} in samples generated by a generator trained with vanilla $\mathcal{L}_{MLE}$ objective --
given a generated sample $(\tilde{P}, \tilde{H}, \tilde{l})$, the label $\tilde{l}$ often does not correctly describe the relationship between $\tilde{P}$ and $\tilde{H}$.
To alleviate this issue, we apply \emph{unlikelihood training}~\citep{unlikelihood} to make generating such label inconsistent instances less likely.
% \ps{Elaborate very briefly why unlikehood is a good pick.}
%

First we perturb the label to construct negative samples $(P, H, l')$ where $l' \neq l$ for each sample in the dataset.
% Then given the premise and perturbed label $x'_l$, 
Then we apply a token-level unlikelihood objective on the hypothesis tokens:
\begin{align*}
    &\mathcal{L}_{\text{consistency}}=  \\
    & -\sum_{i=1}^{|\mathcal{D}_0|} \sum_{t=1}^{|H|^{(i)}} \log (1-p( H_{t}^{(i)}| l'^{(i)}, P^{(i)}, H_{<t}^{(i)})).
\end{align*}
This objective decreases the probability of generating $H$ when given an incorrect label $l'$, hence improves the label consistency at generation time.

We combine $\mathcal{L}_{MLE}$ and $\mathcal{L}_{\text{consistency}}$ to finetune our generator $G$ with 
\begin{equation*}
    \mathcal{L}_{G} = \mathcal{L}_{MLE} + \lambda \mathcal{L}_{\text{consistency}},
\end{equation*}
where $\lambda$ is a hyperparameter that balances the two objectives.
% \footnote{We use $\lambda=0.5$ in our experiments.}
We can randomly sample from the trained generator to obtain a large amount of the synthetic data $\mathcal{D}_{G} \sim G$.

\subsubsection{Filtering Based on Model Confidence} \label{sec:filter}

We add a consistency filtering step~\citep{lewis-etal-2021-paq,bartolo2021improving} to further improve the quality of the generated dataset. We train an NLI model $M$ with the original dataset $\mathcal{D}_0$ to filter out samples in which $M$ has low confidence:
% 
% For sample $(\tilde{P}, \tilde{H}, \tilde{l}) \in \mathcal{D}_{G}$, we filter out the ones that $p_M(\tilde{l}| \tilde{P}, \tilde{H}) < \tau$
\begin{equation*}
    \hat{\mathcal{D}}_{G} = \{(P,H,l) \in \mathcal{D}_{G} \mid p_M(l|P,H) > \tau \},
\end{equation*}
where $\tau$ is a confidence threshold.
We found that the filtered out data samples generally had ungrammatical text or incorrect labels.

% \ps{
%     Motivate \emph{why} we expect this to work.
% }
% \ps{
%     Cite something like Max and PAQ here for filtering?
% }

% =====================================

\section{Mitigating Spurious Correlations using \textit{z-filtering}}\label{sec:fixing_distribution}

We now define a method to reject samples that contribute to the high spurious correlations between task-independent features of the samples and their labels. 
Our approach is based on the theoretical framework proposed by~\citet{competency} to measure these correlations, known as \textit{z-statistics}. Our filtering method, called \textit{z-filtering} (\cref{sec:z-filter}), will serve as the basis to construct debiased datasets in~\cref{sec:construct-dataset}.

\subsection{Identifying and Measuring Spurious Correlations} \label{sec:z-score}

As a first step towards addressing spurious correlations, we need to be able to quantify them.
%quantify spurious correlations before we can mitigate them.
% defining bias features
% We first select a set of simple features that we want our task to be independent with.
We start by selecting a set of task-independent features -- features that give away the labels and allow models to exploit them without actually solving the task.
% We seek to ensure that the labels are not correlated with these features.
% 
For NLI, we choose the following features: %unigrams, bigrams, hypothesis length, hypothesis-premise length ratio, lexical overlap, and a hypothesis-only model's prediction.
\begin{inparaenum}[\itshape 1\upshape)]
    \item unigrams and bigrams;
    \item hypothesis length and hypothesis-premise length ratio;
    \item lexical overlap between hypothesis and premise;
    \item the predictions of a BERT-base~\citep{bert} hypothesis-only model.%
\end{inparaenum}%
\footnote{See \cref{appendix:bias-features} for detailed descriptions of the features.}
These features capture various biases identified in prior work, including contradiction word biases, lexical overlap bias~\citep{hans}, and hypothesis-only bias~\citep{GururanganSLSBS18,PoliakNHRD18}.
Note that our method does not rely on the specific choice of features, and one can easily add alternative features that should not be correlated with the labels.

Following~\citet{competency}, we assume there should be no correlation between each of these features and the class labels.
More formally, for any feature $x$ from our feature set $\mathcal{X}$, $p(l|x)$ should be uniform over the class labels $l$.
We define $\hat{p}(l|x)=\frac{1}{n} \sum_{j=1}^{n}l^j$ to be the empirical expectation of $p(l|x)$ over $n$ samples containing $x$.
Then we compute the standardised version of \textit{z-statistics} to quantify its deviation from the uniform distribution for each feature $x$ and label $l$:
\begin{equation} \label{eq:z-score}
    z^{*}(x,l) = \frac{\hat{p}(l|x)-p_0}{\sqrt{p_0(1-p_0)/n}},
\end{equation}
where $p_0$ is the probability of uniform distribution~($p_0=1/3$ in NLI tasks with three labels).
% advantage of z-statistics
% Z-statistics gives a more complete picture of artifacts than prior pointwise mutual information (PMI) analysis because it also considers the number of times a feature is seen~\citep{competency}.
 
% For any given dataset $\mathcal{D}$, we can calculate the \textit{z-statistic} $z^{*}(x,l | \mathcal{D})$ between any feature $x$ and label $l$. 
% 
These  z-statistics scores can be used to identify the most biased features for each label $l$ -- we select $k$ features with the highest \textit{z-statistic} to define the \emph{biased features} set $\mathcal{B}_{\mathcal{D}}(l)$.
% \begin{equation*}
%   \mathcal{B}_{\mathcal{D}}(l)=\text{top-k}_{x\in \mathcal{X}} z^{*}(x,l|\mathcal{D}).
% \end{equation*}
% 
\cref{tab:biased-features} shows examples of these biased features on SNLI.

\subsection{z-filtering} \label{sec:z-filter}

To mitigate the biases in the dataset,
we propose \emph{z-filtering}, an algorithm that iteratively selects and filters instances from a dataset $\mathcal{D}'$ to build a debiased dataset $\mathcal{Z}$. % (see Algorithm~\ref{alg:z-filter}).
At each step, we find the set of biased features $\mathcal{B}_{\mathcal{Z}}(l)$ on the partially constructed $\mathcal{Z}$.
We then select a new batch of samples from $\mathcal{D}'$ and filter out the samples that contain these biased features.
This process is applied iteratively until it has exhausted all samples from $\mathcal{D}'$.
It removes the samples that contribute to the spurious correlations in $\mathcal{D}'$, thus it finds a debiased subset $\mathcal{Z}(\mathcal{D}') \subset \mathcal{D}'$.
We denote the removed samples as $\mathcal{Z}^{-}(\mathcal{D}')$.
The full z-filtering algorithm is illustrated in Algorithm~\ref{alg:z-filter}.
 
% Z-augment
Optionally, one can initialise $\mathcal{Z}$ with a seed dataset $\mathcal{D}_{seed}$.
In this case, the samples from $\mathcal{D}'$ are only added to $\mathcal{Z}$ when they do not contain the biased features of $\mathcal{D}_{seed}$.
% In practice, this often leads to further reduction of  z-statistics values of the biased features as it increases the denominator $n$ in \cref{eq:z-score}.
Thus it can be seen as a data-augmentation technique targeted to debias a given dataset.
We refer to it as \textit{conditional z-filtering} and denote the produced debiased dataset as $\mathcal{Z}(\mathcal{D}' | \mathcal{D}_{seed})$.

\begin{algorithm}[hbt!]
\caption{z-filtering algorithm.}\label{alg:z-filter}
\KwData{input dataset $\mathcal{D}'$ [with optional seed dataset $\mathcal{D}_{seed}$]}
\KwResult{debiased dataset $\mathcal{Z}$ and the rejected samples $\mathcal{Z}^-$}
$\mathcal{Z} \gets \emptyset$ (or $\mathcal{Z} \gets \mathcal{D}_{seed}$)\;
$\mathcal{Z}^{-} \gets \emptyset$\;
\For{sample batch $\mathcal{D}'_{t} \subset \mathcal{D}'$}
{
    compute or update z-statistics $z^{*}(x,l | \mathcal{Z}), \forall x \in \mathcal{X}$ of $\mathcal{Z}$\;
    find the biased features $\mathcal{B}_{\mathcal{Z}}(l), \forall l \in \{\text{entailment}, \text{neutral}, \text{contradiction}\}$\;
    \ForEach{instance $I=(P, H, l) \in \mathcal{D}'_{t}$}
    {
        get the features $f$ of the instance $I$\;
        \eIf{$f \cap \mathcal{B}_{\mathcal{Z}}(l) = \emptyset$}{
            $\mathcal{Z} \gets \mathcal{Z} \cup \{I\}$\;
        }{
            $\mathcal{Z}^{-} \gets \mathcal{Z}^{-} \cup \{I\}$\;
        }
    }
}
\end{algorithm}

% \ps{
%     At the point of introducing he algorithm we have no clue what this seed dataset is.
% }

\section{Constructing Debiased \ourtask{} Datasets via Data Generation} \label{sec:construct-dataset}

We use z-filtering in two ways:
\begin{inparaenum}[\itshape 1\upshape)]
    \item to further finetune $G$ (the one trained in \cref{sec:genv1_ul} with consistency unlikelihood) with an objective that downweighs samples that should be rejected~(\cref{sec:ul-debias});
    \item to post-hoc filter the generated samples to obtain debiased datasets~(\cref{sec:z-augment}).
\end{inparaenum}

\subsection{Learning to Generate Unbiased Samples} \label{sec:ul-debias}

The generator $G$ can learn to exploit task-independent features during its finetuning stage (\cref{sec:generator}), causing the synthetic data $\hat{\mathcal{D}}_{G}$ to contain many spurious correlations.
While it is tempting to apply z-filtering to remove these spurious correlations from $\hat{\mathcal{D}}_{G}$, we find that this will lead to the removal of majority of the generated data.
For example, when the generator is finetuned on SNLI, z-filtering removes around 85\% of $\hat{\mathcal{D}}_{G_{SNLI}}$.%
\footnote{This is also strong confirmation that these biases are problematic, as the generative model easily finds them and relies on them during data generation. Conducting naive data augmentation with $\hat{\mathcal{D}}_{G_{SNLI}}$ will strengthen the spurious correlations.}
% \matt{Consider adding a footnote that says something like ``This is also strong confirmation that these biases are problematic, as the generative model easily finds them and relies on them during data generation.''}
%
This leads to a very inefficient data generation process to mitigate the spurious correlations.

To alleviate this issue, we can incorporate the debiasing objectives into the training of the generator, so that the samples produced by the generator are more likely to be accepted by the z-filtering process.
More specifically, we can encourage the model to generate $\mathcal{Z}(\mathcal{D}_0)$, while discouraging it from generating $\mathcal{Z}^-(\mathcal{D}_0)$.
For the latter part, we again apply an unlikelihood training objective $\mathcal{L}_{UL}$ to \emph{unlearn} $\mathcal{Z}^-(\mathcal{D}_0)$. 
% \matt{``degenerate'' is not a verb. I think you're looking for a different term here.}
% 
Hence, the overall debiasing training objective is:
\begin{equation*}
\mathcal{L}_{debias} = \mathcal{L}_{MLE}(\mathcal{Z}(\mathcal{D}_0)) + \alpha  \mathcal{L}_{UL}(\mathcal{Z}^-(\mathcal{D}_0))
\end{equation*}
where $\alpha$ is a hyperparameter. 

A naive use of an unlikelihood objective on all tokens gives the model mixed signals for good tokens and leads to ungrammatical, degenerate outputs. 
To avoid this degeneracy, we apply the unlikelihood loss only to tokens that contribute to biased features.
% The unlikelihood loss applies only to the tokens that contribute to biased features.
Concretely, for each token $I^-_t$ of instance $I^- \in \mathcal{Z}^-(\mathcal{D}_0)$, we define a mask $m_t$ as
\begin{equation*}
    m_t = 
\begin{cases}
    0,& \text{if } I^-_t \text{ contributes to }\mathcal{B}_\mathcal{Z}(l_{I^-})\\
    1,              & \text{otherwise}.
\end{cases}
\end{equation*}
where $\mathcal{B}_\mathcal{Z}(l_{I^-})$ represent the biased features corresponding the label of $I^-$.

For biases towards unigram and bigram features (as defined in~\cref{sec:z-score}), we consider only the corresponding tokens to be relevant (i.e., $m_t = 0$ if $I_t^-$ is part of the unigram or the bigram). For biases towards other features (e.g. length of the hypothesis), we consider all the tokens on the hypothesis to be relevant. The unlikelihood training objective is defined as follows:
\begin{align*}
\mathcal{L}_{UL}(\mathcal{Z}^-(\mathcal{D}_0))=&\sum_{I' \in \mathcal{Z}^-(\mathcal{D}_0)} \mathcal{L}_{UL}(I'), \\
\mathcal{L}_{UL}(I')=- \sum_{t=1}^{|I'|} \log ( & m_t p( I'_{t}| I'_{<t}) \\
     + &(1-m_t) (1-p( I'_{t}| I'_{<t})) ).
\end{align*}

We further finetune $G$ with $\mathcal{L}_{debias}$ to obtain a new generator $G^*$, that is trained to generate more unbiased data samples.
We then randomly sample from $G^*$ and conduct data filtering~(\cref{sec:filter}) to obtain a large set of high-quality debiased data samples $\hat{\mathcal{D}}_{G^*}$.

\subsection{Combining with z-filtering to Construct the Debiased NLI Datasets} \label{sec:z-augment}

Given the original dataset $\mathcal{D}_0$ and the synthetic dataset $\hat{\mathcal{D}}_{G^*}$, our goal is produce a large-scale unbiased dataset $\mathcal{D}^{*}$.
There are various ways to do this given that we can either apply conditional z-filtering, or simply z-filter both $\mathcal{D}_0$ and $\hat{\mathcal{D}}_{G^*}$ and merge them.
We explore the following options:
\begin{enumerate}
    \item \textbf{Z-Augmentation (Z-Aug)} $\mathcal{Z}(\hat{\mathcal{D}}_{G^*} | \mathcal{D}_0)$: we keep the original dataset as is, and augment it by conducting conditional z-filtering on $\hat{\mathcal{D}}_{G^*}$ using $\mathcal{D}_0$ as seed dataset.
    \item \textbf{Parallel z-filter (Par-Z)} $\mathcal{Z}(\mathcal{D}_0) \cup \mathcal{Z}(\hat{\mathcal{D}}_{G^*})$: we conduct z-filtering on $\mathcal{D}_0$ and $\hat{\mathcal{D}}_{G^*}$ separately, and then merge them.
    \item \textbf{Sequential z-filter (Seq-Z)} $\mathcal{Z}( \hat{\mathcal{D}}_{G^*} | \mathcal{Z}(\mathcal{D}_0))$: we first conduct z-filtering on $\mathcal{D}_0$, then conduct conditional z-filtering on $\hat{\mathcal{D}}_{G^*}$ with $\mathcal{Z}(\mathcal{D}_0)$ as seed dataset.
\end{enumerate}

% \ps{
%     I think this part is best served with an illustration as well. If the reader does not ``get'' this part, we are in somewhat of a pickle.
% }

%================= Experiments =================
\section{Experiments} \label{sec:exp}
\subsection{Experimental Setup}

\paragraph{Source Datasets}

We select the two most widely used \ourtask{} datasets SNLI~\citep{snli} and MNLI~\citep{mnli} as our original datasets.
Prior work~\citep{GururanganSLSBS18,PoliakNHRD18,hans} found various annotation artifacts in them, hence they serve as good use cases for constructing debiased datasets.

\paragraph{Evaluation Datasets}
% follow-up work proposed several benchmarks to evaluate the robustness against these biases.
% 
For the hypothesis-only bias, we use the challenge sets SNLI-hard~\citep{GururanganSLSBS18} and MNLI-hard~\citep{mnli}, which were produced by filtering the test set with a hypothesis-only model~(\cref{sec:nli-hard}).
For syntactic biases, we follow previous work and use HANS~\citep{hans} for evaluation~(\cref{sec:hans-results}).
In addition, we evaluate on the adversarial test benchmark introduced by~\citet{model-agnostic-debias}~(\cref{sec:nli-adv-test}).
This benchmark covers a wide range of adversarial attacks, which will give a more complete picture of what spurious correlations the debiasing methods tackle. 
% \matt{finish this sentence}

\paragraph{Generating Debiased Datasets}
We conduct debiased data generation for SNLI and MNLI \emph{separately}.
For SNLI, we use the proposed method described in \cref{sec:ul-debias} to train a generator $G^*_{\text{SNLI}}$.
Then we randomly sample a large number of instances from the generator to construct $\mathcal{D}_{G^*_{\text{SNLI}}}$.
The samples are filtered with a strong NLI model $M$ trained on SNLI to obtain  $\hat{\mathcal{D}}_{G^*_{\text{SNLI}}}$.
Finally, different options (\cref{sec:z-augment}) can be adopted to merge the synthetic data with the original data $\mathcal{D}_{\text{SNLI}}$  to construct debiased versions of SNLI.
The same procedure is used to produce debiased datasets for MNLI, by simply replacing the original dataset with MNLI.
We choose GPT-2 large and Roberta-large as the pretrained language models for $G^*$ and $M$ respectively.%
\footnote{On one A100 GPU, training the generator takes around 24 hours and generating the samples takes roughly 35 hours for each dataset.}
% and $N=5$ million on SNLI and $N=4$ million on MNLI.%
% \ps{
%     What is the motivation for these parameters? They currently seem like magic to me.
% }
% 
The size of the constructed debiased datasets are listed in~\cref{tab:data-size}.

\begin{table}[bt]
\begin{center}
\resizebox{\columnwidth}{!}{
    \begin{tabular}{lrr}
      \toprule
      {\bf Options} & {\bf $\mathcal{D}_0=\mathcal{D}_{\text{SNLI}}$  } & {\bf $\mathcal{D}_0=\mathcal{D}_{\text{MNLI}}$ } \\
      \midrule
      \textbf{Original} $\mathcal{D}_0$  & 549,367 &  382,702 \\
      \textbf{Z-Aug} $\mathcal{Z}(\hat{\mathcal{D}}_{G^*} | \mathcal{D}_0)$ &  1,142,475 & 744,326 \\
      \textbf{Par-Z} $\mathcal{Z}(\mathcal{D}_0) \cup \mathcal{Z}(\hat{\mathcal{D}}_{G^*})$ & 933,085 & 740,811 \\
      \textbf{Seq-Z} $\mathcal{Z}( \hat{\mathcal{D}}_{G^*} | \mathcal{Z}(\mathcal{D}_0))$ & 927,906 & 744,200 \\
      \bottomrule
    \end{tabular}
}
\caption{Data size of the constructed debiased datasets for SNLI and MNLI.} \label{tab:data-size}
\end{center}
\end{table}

\paragraph{NLI Model Training}

Since our method directly debiases the training data itself, we keep the model and training objective fixed and only replace the training data with our generated debiased datasets.
For comparability with previous work~\citep{karimi-mahabadi-etal-2020-end,utama-etal-2020-mind,sanh2021learning}, we train BERT-base~\citep{bert} on our debiased datasets.
The NLI models are trained with ordinary \emph{cross-entropy} classification loss,
and the training hyperparameters are listed in~\cref{appendix:hparams}.
We run our experiments five times and report the average and standard deviation of the scores.\footnote{With the exception of our PoE experiments which single run, as hyperparameter tuning for PoE is costlier.}
We also conduct statistical significance testing using a 2-tailed t-test at 95\% confidence level.

\paragraph{State-of-the-art Debiasing Models}
We compare our method with the following three state-of-the-art debiasing models on each of our evaluation datasets.
\textbf{Product-of-Experts}~\citep{he-etal-2019-unlearn,karimi-mahabadi-etal-2020-end} ensembles a bias-only model's prediction $b_i$ with the main model's $p_i$ using $p'_i=softmax(\log p_i + \log b_i)$.
This ensembling enforces that the main model focuses on the samples that the bias-only model does not predict well.
\textbf{Learned-Mixin}~\citep{clark-etal-2019-dont} is a variant of PoE that introduces a learnable weight for the bias-only model's prediction.
\textbf{Regularized-conf}~\citep{utama-etal-2020-mind} uses confidence regularisation to retain the in-distribution performance while conducting model debiasing.

\paragraph{Combining PoE with Our Debiased Datasets}
Our approach changes the training data distribution instead of the model's training objective, and hence is orthogonal to prior work method-wise.
We also report the results of combining PoE with our proposed method, simply by training a PoE model on our debiased datasets.
We adapt the PoE implementation by~\citet{karimi-mahabadi-etal-2020-end}, and we follow their approach to conduct hyperparameter tuning for PoE.\footnote{\url{https://github.com/rabeehk/robust-nli}}
The hyperparameters of the PoE models are reported in \cref{tab:poe-hparams} of \cref{appendix:hparams}.

\begin{table}[bt]
\begin{center}
\resizebox{\columnwidth}{!}{
    \begin{tabular}{lcl}
      \toprule
      {\bf Method (model w/ data)} & {\bf SNLI} & {\bf SNLI-hard} \\
      \midrule
      \multicolumn{3}{l}{\bf{Prior debiasing strategies trained on SNLI}} \\
      {AdvCls~\citep{belinkov-etal-2019-dont}}$*$ & 83.56 & 66.27\\
      {Ens. AdvCls~\citep{stacey-etal-2020-avoiding}}$*$ & 84.09 & 67.42  \\
      {DFL~\citep{karimi-mahabadi-etal-2020-end}}$*$ & 89.57 & \textbf{83.01} \\
      {PoE~\citep{karimi-mahabadi-etal-2020-end}}$*$ & 90.11 & 82.15  \\
      \midrule
      BERT-base w/ $\mathcal{D}_{\text{SNLI}}$ \emph{baseline} & 90.45 & 80.34$_{\pm 0.46}$ \\
      \multicolumn{3}{l}{\bf{Models trained on our debiased datasets}} \\
      BERT-base w/ Z-Aug $\mathcal{Z}(\hat{\mathcal{D}}_{G^*} | \mathcal{D}_{\text{SNLI}})$ & \textbf{90.67} & \underline{81.78}$_{\pm 0.53}$ \\
      BERT-base w/ Par-Z $\mathcal{Z}(\mathcal{D}_{\text{SNLI}}) \cup \mathcal{Z}(\hat{\mathcal{D}}_{G^*})$ & 88.11 & \underline{82.81}$_{\pm 0.37}$ \\
      BERT-base w/ Seq-Z $\mathcal{Z}( \hat{\mathcal{D}}_{G^*} | \mathcal{Z}(\mathcal{D}_{\text{SNLI}}))$ & 88.08 & \underline{\textbf{82.82}}$_{\pm 0.15}$ \\
      \midrule
      \multicolumn{3}{l}{\bf{Combining PoE with our debiased datasets}} \\
      BERT-base + PoE w/ $\mathcal{D}_{\text{SNLI}}$ & 90.25 & 82.92 \\
      BERT-base + PoE w/ Seq-Z $\mathcal{Z}( \hat{\mathcal{D}}_{G^*} | \mathcal{Z}(\mathcal{D}_{\text{SNLI}}))$ & 87.65 & \textbf{84.48} \\
      \bottomrule
    \end{tabular}
}
\caption{Accuracy on SNLI and SNLI-hard. $*$ are reported results and underscore indicates statistical significance against the baseline.
% with t-test at 95\% confidence level. 
Training on our debiased SNLI datasets significantly boosts the performance on SNLI-hard compared to the baseline, and it improves further when combined with PoE.} \label{tab:snli-results}
\end{center}
\end{table}

\begin{table*}[bt]
\begin{center}
\resizebox{\textwidth}{!}{
    \begin{tabular}{lllllllll}
      \toprule
      {\bf Method (model w/ data)} &  \multicolumn{2}{c}{\bf{MNLI-m}} &  \multicolumn{2}{c}{\bf{MNLI-mm}} &\multicolumn{2}{c}{\bf{MNLI-m hard}}  &  \multicolumn{2}{c}{\bf{MNLI-mm hard}} \\
       &  \multicolumn{1}{c}{dev} & \multicolumn{1}{c}{test} &  \multicolumn{1}{c}{dev} & \multicolumn{1}{c}{test} &  \multicolumn{1}{c}{dev} & \multicolumn{1}{c}{test} &  \multicolumn{1}{c}{dev} & \multicolumn{1}{c}{test} \\
      \midrule
      \multicolumn{9}{l}{\bf{Prior debiasing strategies trained on MNLI}} \\
      {PoE~\citep{karimi-mahabadi-etal-2020-end}}$*$ & 84.58 & 84.11 & 84.85 & 83.47 & 78.02 & 76.81 & 79.23 & 76.83 \\
      {Learned-Mixin~\citep{clark-etal-2019-dont}}$*$ & 80.5 & 79.5 & 81.2 & 80.4 & - & 79.2 & - & 78.2 \\
      {Regularized-conf~\citep{utama-etal-2020-mind}}$*$ & 84.6 & 84.1 & 85.0 & 84.2 & - & 78.3 & - & 77.3 \\
      {BERT-base Main PoE+CE~\citep{sanh2021learning}}$*$ & 83.32 & - & 83.54 & - & - & 77.63 & - & 76.39 \\
      \midrule
      BERT-base w/ $\mathcal{D}_{\text{MNLI}}$ \textit{baseline} & 83.87 & 84.11 & 84.22 & 83.51 & 76.39$_{\pm 0.64}$ & 75.88 & 77.75$_{\pm 0.45}$ & 75.75 \\
      \multicolumn{9}{l}{\bf{Models trained on our debiased datasets}} \\
      BERT-base w/ Z-Aug $\mathcal{Z}(\hat{\mathcal{D}}_{G^*} | \mathcal{D}_{\text{MNLI}})$ & \textbf{84.72} & \textbf{85.12} &\textbf{85.14} & \textbf{84.09} & \underline{\textbf{78.95}}$_{\pm 0.76}$ & 78.60 & \underline{\textbf{80.29}}$_{\pm 0.54}$ & \textbf{78.51} \\
      BERT-base w/ Par-Z $\mathcal{Z}(\mathcal{D}_{\text{MNLI}}) \cup \mathcal{Z}(\hat{\mathcal{D}}_{G^*})$  & 82.48 & 83.27 & 82.95 & 82.95 & \underline{78.88}$_{\pm 0.80}$ & \textbf{79.19} & \underline{80.02}$_{\pm 0.62}$ & 78.49 \\
      BERT-base w/ Seq-Z $\mathcal{Z}( \hat{\mathcal{D}}_{G^*} | \mathcal{Z}(\mathcal{D}_{\text{MNLI}}))$  & 82.55 & 83.41 & 82.70 & 83.17 & \underline{78.88}$_{\pm 0.83}$ & \textbf{79.19} & \underline{79.65}$_{\pm 0.44}$ & 78.44 \\
      \midrule
      \multicolumn{9}{l}{\bf{Combining PoE with our debiased dataset}} \\
      BERT-base + PoE w/ $\mathcal{D}_{\text{MNLI}}$ & 84.39 & 84.69 & 84.25 & 83.75 & 78.37 & 77.54 & 79.45 & 78.33 \\
      BERT-base + PoE w/ Z-Aug $\mathcal{Z}(\hat{\mathcal{D}}_{G^*} | \mathcal{D}_{\text{MNLI}})$ & \textbf{85.22} & \textbf{85.38} & \textbf{85.72}& \textbf{84.53} & \textbf{80.49} & \textbf{80.03} & \textbf{81.52} & \textbf{79.28} \\
      \bottomrule
    \end{tabular}
}
\caption{Accuracy on MNLI-matched (MNLI-m), MNLI-mismatched (MNLI-mm), MNLI-matched hard, and MNLI-mismatched hard. $*$ are reported results and underscore indicates statistical significance against the baseline.
% with t-test at 95\% confidence level. 
Training on our debiased MNLI datasets significantly boosts the performance on MNLI-matched hard and MNLI-mismatched hard. When combined with PoE, our method improves further and outperforms previous methods.
} \label{tab:mnli-results}
%\matt{Looks like MNLI-mm test results have incorrect highlighting}
\end{center}
\end{table*}

\subsection{Hypothesis-only Bias in NLI} \label{sec:nli-hard}

\citet{GururanganSLSBS18} found that, on SNLI and MNLI, a model that only has access to the hypothesis can perform surprisingly well, which indicates that the datasets contain hypothesis-only bias.
To alleviate this problem, SNLI-hard and MNLI-hard~\citep{GururanganSLSBS18} subsets were constructed by filtering the test set with a hypothesis-only model and only accepting those that the hypothesis-only model predicts incorrectly.
We examine whether our method successfully mitigates the hypothesis-only bias in NLI, by evaluating the models trained with our debiased datasets on SNLI-hard and MNLI-hard.

\begin{table}[!bt]
\begin{center}
\resizebox{\columnwidth}{!}{
    \begin{tabular}{ll}
      \toprule
      {\bf Method}  & {\bf HANS} \\
      \midrule
      \midrule
      \multicolumn{2}{l}{\bf{Methods trained on SNLI}} \\
      \midrule
      {BERT-base Attention~\citep{joe-explain}}$*$ & 58.42 \\
      {Roberta-large w/ AFLite~\citep{AFLite}}$*$ & 59.6 \\
    %   {Roberta-base w/ $\mathcal{D}_{\text{SNLI}}$~\citep{ross2021tailor}}$*$ & 64.72 \\
      {Roberta-base w/ TAILOR~\citep{ross2021tailor}}$*$ & 70.5 \\
      \hline \hline
      \multicolumn{2}{l}{\bf{Methods trained on MNLI}} \\
      \midrule
      {Learned-Mixin~\citep{clark-etal-2019-dont}}$*$ &  64.00 \\
      {Learned-Mixin+H~\citep{clark-etal-2019-dont}}$*$ &  66.15 \\
      {PoE~\citep{karimi-mahabadi-etal-2020-end}}$*$ &  66.31$_{\pm 0.6}$ \\
      {DFL~\citep{karimi-mahabadi-etal-2020-end}}$*$ &  69.26$_{\pm 0.2}$ \\
      {PoE+CE~\citep{sanh2021learning}}$*$  & 67.9 \\
      {Regularized-conf~\citep{utama-etal-2020-mind}}$*$ &  69.1$_{\pm 1.2}$ \\
      {E2E Self-debias~\citep{ghaddar-etal-2021-end}}$*$ & \textbf{71.2}$_{\pm 0.2}$ \\
      \hline \hline
      \multicolumn{2}{l}{\bf{Models trained on our debiased datasets}} \\
      \midrule
      Roberta-base w/ $\mathcal{D}_{\text{SNLI}}$ & 65.32$_{\pm 2.22}$\\
      Roberta-base w/ Seq-Z $\mathcal{Z}( \hat{\mathcal{D}}_{G^*} | \mathcal{Z}(\mathcal{D}_{\text{SNLI}}))$ & \textbf{66.87}$_{\pm 1.47}$ \\
      \midrule
      BERT-base w/ $\mathcal{D}_{\text{MNLI}}$ \emph{baseline} & 54.36$_{\pm 2.56}$ \\
      BERT-base w/ Z-Aug $\mathcal{Z}(\hat{\mathcal{D}}_{G^*} | \mathcal{D}_{\text{MNLI}})$ & \underline{62.57}$_{\pm 5.91}$  \\
      BERT-base w/ Par-Z $\mathcal{Z}(\mathcal{D}_{\text{MNLI}}) \cup \mathcal{Z}(\hat{\mathcal{D}}_{G^*})$ & \underline{65.11}$_{\pm 5.62}$  \\
      BERT-base w/ Seq-Z $\mathcal{Z}( \hat{\mathcal{D}}_{G^*} | \mathcal{Z}(\mathcal{D}_{\text{MNLI}}))$ & \underline{\textbf{67.69}}$_{\pm 3.53}$   \\
      \midrule
      BERT-base + PoE w/ $\mathcal{D}_{\text{MNLI}}$ (baseline) & 63.40 \\
      BERT-base + PoE w/ Z-Aug $\mathcal{Z}(\hat{\mathcal{D}}_{G^*} | \mathcal{D}_{\text{MNLI}})$ & \textbf{68.75} \\
      \midrule
      Roberta-large w/ $\mathcal{D}_{\text{MNLI}}$ & 75.74$_{\pm 2.82}$ \\
      Roberta-large w/ Z-Aug $\mathcal{Z}(\hat{\mathcal{D}}_{G^*} | \mathcal{D}_{\text{MNLI}})$ & \textbf{78.65}$_{\pm 2.26}$  \\
      \bottomrule
    \end{tabular}
}
\caption{Results on HANS~\citep{hans}. $*$ are reported results and underscore indicates statistical significance against the baseline.
BERT-base trained on our debiased MNLI datasets performs significantly better than the one trained on the original MNLI, and it improves further when combined with PoE. Roberta-large also benefits from training on our debiased dataset.
} \label{tab:hans-results}
\end{center}
\end{table}

\paragraph{Results on SNLI-hard}
\cref{tab:snli-results} shows the results of our method on SNLI and SNLI-hard. 
The results show that, compared to training on SNLI, training with our debiased datasets significantly improves the performance on SNLI-hard. 
The debiased dataset produced by Seq-Z achieves a 2.48\% gain in accuracy on SNLI-hard compared to the SNLI baseline, 
whereas Z-Aug improves both SNLI and SNLI-hard accuracy.
% 
% Seq-Z's performance drop on SNLI indicates that conducting z-filtering on SNLI effectively removes the biased samples, but it also affects the in-distribution performance negatively.
% 
% Z-Aug retains all SNLI samples (which includes biased samples), thus it is hard to mitigate all spurious correlations.
% 

\paragraph{Results on MNLI-hard}
\cref{tab:mnli-results} shows the results of our method on MNLI-matched (MNLI-m) and MNLI-mismatched (MNLI-mm), and their corresponding hard sets.
We use the development sets of MNLI-hard reconstructed by~\citep{karimi-mahabadi-etal-2020-end} to develop our methods. 
To comply with the submission limit of MNLI leaderboard system,
% \footnote{\tiny{\url{https://www.kaggle.com/c/multinli-mismatched-open-hard-evaluation}}}
we select the best checkpoint among the five runs using the development set, and report its test set performance in~\cref{tab:mnli-results}.

The results show that BERT-base models trained on our debiased MNLI datasets outperform the models trained on the original MNLI by a large margin on the MNLI-hard sets.
In particular, the Z-Aug version of the debiased datasets gives a 2.72\% and 2.76\% gain in accuracy on MNLI-m hard and MNLI-mm hard respectively, and outperforms the previous state-of-the-art on MNLI-m, MNLI-mm, and MNLI-mm hard.
% 
% The other two debiased versions obtain similar performance gains on the hard sets, but we observe a similar drop of in-distribution accuracy, which is akin to the results on SNLI.

\paragraph{Combining PoE with Our Debiased Datasets}

We investigate the combination of our method and PoE, to see if the two orthogonal techniques can work together to achieve better performance.
Since hyperparameter tuning of PoE is costly, we choose the best version of the debiased dataset~(Seq-Z for SNLI and Z-Aug for MNLI) using the development set accuracy, and train PoE with it.
The results are listed in the last rows of \cref{tab:snli-results} and \cref{tab:mnli-results}.
We can find that, on both SNLI and MNLI, combining PoE with our debiased dataset yields further improvements on SNLI-hard, MNLI-m hard, and MNLI-mm hard, outperforming previous state-of-the-art results on all three datasets.
% 
% \yw{With the results on SNLI and MNLI, we can find that the debiased datasets generated by our proposed method can effectively improve ...}
% \yw{Discussion of why this is happening}

\begin{table*}[!bt]
\begin{center}
\resizebox{\textwidth}{!}{
    \begin{tabular}{lllllllll}
      \toprule
    %   {\bf Method (model w/ data)} &  \multicolumn{2}{c}{\bf{MNLI-m}} &  \multicolumn{2}{c}{\bf{MNLI-mm}} &\multicolumn{2}{c}{\bf{MNLI-m hard}}  &  \multicolumn{2}{c}{\bf{MNLI-mm hard}} \\
       &  \textbf{PI-CD} & \textbf{PI-SP} & \textbf{IS-SD} & \textbf{IS-CS} & \textbf{LI-LI} & \textbf{LI-TS} & \textbf{ST} & \textbf{Avg.} \\
      \midrule
      \multicolumn{9}{l}{\bf{Data-augmentation heuristics proposed by~\citet{model-agnostic-debias}}} \\
      {Text Swap}$*$ & 71.7 & 72.8 & 63.5 & 67.4 & 86.3 & \textbf{86.8} & 66.5 & 73.6 \\
      {Sub (synonym)}$*$ & 69.8 & 72.0 & 62.4 & 65.8 & 85.2 & 82.8 & 64.3 & 71.8 \\
      {Sub (MLM)}$*$ & 71.0 & 72.8 & 64.4 & 65.9 & 85.6 & 83.3 & 64.9 & 72.6 \\
      {Paraphrase}$*$ & 72.1 & 74.6 & 66.5 & 66.4 & 85.7 & 83.1 & 64.8 & 73.3 \\
      \midrule
      BERT-base w/ $\mathcal{D}_{\text{MNLI}}$ \textit{baseline} & 70.3$_{\pm 0.5}$ & 73.7$_{\pm 1.4}$ & 53.5$_{\pm 2.3}$ & 64.8$_{\pm 1.4}$ & 85.5$_{\pm 0.9}$ & 81.6$_{\pm 1.4}$ & 69.2$_{\pm 0.8}$ & 71.2$_{\pm 0.8}$ \\
      \multicolumn{9}{l}{\bf{Models trained on our debiased datasets}} \\
      BERT-base w/ Z-Aug $\mathcal{Z}(\hat{\mathcal{D}}_{G^*} | \mathcal{D}_{\text{MNLI}})$  & \underline{\textbf{73.1}}$_{\pm 0.9}$ & \underline{76.1}$_{\pm 1.2}$ & \underline{61.8}$_{\pm 6.1}$ & \underline{69.1}$_{\pm 1.3}$ & \underline{86.9}$_{\pm 0.6}$ & 83.1$_{\pm 0.9}$ & \textbf{70.1}$_{\pm 0.5}$ & \underline{74.3}$_{\pm 1.3}$ \\
      BERT-base w/ Par-Z $\mathcal{Z}(\mathcal{D}_{\text{MNLI}}) \cup \mathcal{Z}(\hat{\mathcal{D}}_{G^*})$  & \underline{72.0}$_{\pm 0.9}$ & \underline{\textbf{78.7}}$_{\pm 1.2}$ & \underline{64.5}$_{\pm 5.8}$ & \underline{70.7}$_{\pm 1.7}$ & \underline{88.5}$_{\pm 0.7}$ & 82.6$_{\pm 0.3}$ & 69.6$_{\pm 1.0}$ & \underline{75.2}$_{\pm 1.4}$ \\
      BERT-base w/ Seq-Z $\mathcal{Z}(\hat{\mathcal{D}}_{G^*} | \mathcal{Z}(\mathcal{D}_{\text{MNLI}}))$ & \underline{71.7}$_{\pm 0.9}$ & \underline{77.8}$_{\pm 1.2}$ & \underline{\textbf{66.9}}$_{\pm 3.7}$ & \underline{\textbf{71.1}}$_{\pm 0.7}$ & \underline{\textbf{89.1}}$_{\pm 1.0}$ & 82.3$_{\pm 0.9}$ & 69.3$_{\pm 0.8}$ & \underline{\textbf{75.4}}$_{\pm 0.8}$ \\
    %   \midrule
    %   \multicolumn{9}{l}{\bf{Combining PoE with our debiased dataset}} \\
    %   BERT-base + PoE w/ $\mathcal{Z}(\hat{\mathcal{D}}_{G^*} | \mathcal{D}_{\text{MNLI}})$ & \textbf{85.22} & \textbf{85.38} & \textbf{85.72}& \textbf{84.53} & \textbf{80.49} & \textbf{80.03} & \textbf{81.52} & \textbf{79.28}  \\
      \bottomrule
    \end{tabular}
}
\caption{Results on the NLI adversarial test benchmark~\citep{model-agnostic-debias}. We compare with the data augmentation techniques investigated by~\citet{model-agnostic-debias}. % and $*$ are reported results.
$*$ are reported results and underscore indicates statistical significance against the baseline.
Training on our debiased MNLI datasets significantly improves the performance on majority of the categories (PI-CD, PI-SP, IS-SD, IS-CS, LI-LI) and on average.
} \label{tab:adv-attack-results}
% \matt{There are a lot of numbers here, not sure you really need all of these rows and/or columns.  This would be an easy place to recover some space if you need to.}
\end{center}
\end{table*}

\subsection{Syntactic Bias in NLI} \label{sec:hans-results}

\citet{hans} show that NLI models trained on MNLI can exploit syntactic heuristics present in the data, such as lexical overlap, subsequence, and constituent features.
They introduce HANS, an evaluation dataset that contains examples where the syntactic heuristics fail.
To test whether our method mitigates the syntactic biases in NLI, we evaluate models trained on our debiased datasets on HANS.
If our debiased dataset contains less syntactic bias than the original dataset, the model would not exploit the syntactic heuristics and thus perform better on HANS.
Due to the high variance of the scores on HANS, we run five times for each experiment (except PoE), and report the average and standard deviation of the scores.

\paragraph{Results on HANS}
\cref{tab:hans-results} shows the results on HANS.
The results are categorised into three sections according to the training data: SNLI, MNLI, and our debiased datasets.
% 
% With SNLI as the original dataset, we compare with TAILOR~\citep{ross2021tailor}, a semantically controlled data augmentation method that uses heuristics specifically designed to tackle syntactic biases.
% 
% Following TAILOR, we train Roberta-base models with our debiased SNLI dataset (Seq-Z).
% 
% The results show that the performance of our debiased dataset outperforms the SNLI baseline, and is also slightly better than TAILOR.
% 
% This is surprising because TAILOR relies on specifically designed heuristics to generate samples, whereas our method does not require such manual heuristics.
% 
The results of models trained on our debiased MNLI datasets show strong improvements: compared to the original MNLI, our debiased MNLI datasets obtain up to a 13.33\% gain in HANS accuracy.
Our Seq-Z variant achieves 67.69\% accuracy, which is comparable with strong PoE baseline~\citep{karimi-mahabadi-etal-2020-end, sanh2021learning}. 
% \matt{Shouldn't you be highlighting the roberta-large result here?  Also, why is MNLI split between MNLI/debiased MNLI while SNLI isn't?}
Our method also further improves PoE models: the BERT-base PoE model trained on our Z-Aug MNLI outperforms the one trained on MNLI by 5.3\%.
Additionally, training Roberta-large~\citep{roberta} on our debiased dataset introduces 2.9 points accuracy gain on HANS, indicating that the performance gain by our debiased dataset can generalise to larger and stronger models (more on this in \cref{sec:larger-models}).

\subsection{Adversarial Tests for Combating Distinct Biases in NLI} \label{sec:nli-adv-test}

\citet{model-agnostic-debias} find that debiasing methods often tie to one particular known bias and it is nontrivial to mitigate multiple NLI biases at the same time.
They introduce a suite of test datasets for NLI models that targets various aspects of robustness, including partial input heuristics (PI), logical inference ability (LI), and stress test (ST).%
\footnote{Details of the subcategories are described in~\cref{appendix:adv-test-desc}.}
Several data augmentation strategies were investigated by~\citet{model-agnostic-debias}:
\begin{inparaenum}[1)]
\item text swap: swapping the premise and hypothesis in the original data;
\item word substitution: replacing words in the hypothesis with synonyms or generations from a masked language model;
\item paraphrase: using back translation to paraphrase the hypothesis.
\end{inparaenum}

We compare our approach with their data-augmentation heuristics, and the results are shown in~\cref{tab:adv-attack-results}.
Comparing with the MNLI baseline, our debiased MNLI datasets lead to better performance across all categories, which indicates that our method successfully mitigates various distinct biases simultaneously.
All three variants of our debiased datasets outperform the data augmentation heuristics by~\citet{liu2021toward},
which demonstrates the efficacy of our method when compared against manually designed heuristics.

\subsection{Generalisation to Larger Pretrained Language Models} \label{sec:larger-models}

Since our method mitigates the spurious correlations in the dataset, not the model, our approach is model-agnostic and has the potential to benefit larger future models.
To test this hypothesis, we train stronger and more modern models than BERT with our debiased datasets, and see if it can still improve the performance.
More specifically, we choose Roberta-base, Roberta-large~\citep{roberta}, and Albert-xxlarge~\citep{albert}, train them with Seq-Z SNLI and Z-Aug MNLI.

The results in \cref{tab:stronger-models} show that: 
\begin{inparaenum}[1)]
\item these larger models achieve better generalisation performance than BERT-base, which agrees with \citet{generalise-nli,combat-hype-caution};
\item training on our debiased datasets can still improve the performance of these models, yielding an average 2.30\%, 1.23\%, 1.13\% gain for Roberta-base, Roberta-large and Albert-xxlarge respectively.
\end{inparaenum}
% Only one run for Albert-xxlarge because it is much costlier to train the model.
This indicates that our method generalises to larger pretrained language models and could potentially enhance future models.

\begin{table}[hbt]
\begin{center}
\resizebox{\columnwidth}{!}{
    \begin{tabular}{llllc}
      \toprule
       & \textbf{Test data} & \textbf{Original} & \textbf{Debiased} & \textbf{$\Delta$} \\
      \midrule
      \multirow{ 5}{*}{\rotatebox[origin=c]{90}{Roberta-base}} & SNLI-hard & 82.02$_{\pm 0.24}$ & \underline{\textbf{83.71}}$_{\pm0.31}$ & 1.69 \\
      & MNLI-m hard & 81.74$_{\pm0.44}$ & \underline{\textbf{83.14}}$_{\pm0.25}$ & 1.40 \\
      & MNLI-mm hard &  81.93$_{\pm0.30}$ & \underline{\textbf{83.12}}$_{\pm0.24}$ & 1.19 \\
      & HANS & 71.17$_{\pm2.95}$ & \underline{\textbf{76.15}}$_{\pm1.52}$ & 4.98 \\
      & Adv.Test avg & 77.63$_{\pm0.49}$ & \underline{\textbf{79.89}}$_{\pm0.38}$ & 2.26 \\
      \midrule
      \multirow{ 5}{*}{\rotatebox[origin=c]{90}{Roberta-large}} & SNLI-hard & 83.61$_{\pm0.31}$ & \underline{\textbf{85.09}}$_{\pm0.32}$ & 1.48 \\
      & MNLI-m hard & 85.44$_{\pm0.62}$ &  \textbf{85.69}$_{\pm0.24}$ &  0.25\\
      & MNLI-mm hard &  85.37$_{\pm0.63}$ & \textbf{85.94}$_{\pm0.21}$  & 0.57 \\
      & HANS & 75.74$_{\pm2.82}$ & \textbf{78.65}$_{\pm2.26}$ & 2.91 \\
      & Adv.Test avg & 80.92$_{\pm0.46}$ & \underline{\textbf{81.86}}$_{\pm0.31}$ & 0.94 \\
      \midrule
      \multirow{ 5}{*}{\rotatebox[origin=c]{90}{Albert-xxlarge}} & SNLI-hard & 83.59& \textbf{84.82} & 1.23 \\
      & MNLI-m hard & \textbf{86.42} & 86.40 & -0.02 \\
      & MNLI-mm hard &  86.38 & \textbf{86.82 }& 0.44 \\
      & HANS & 76.32 & \textbf{79.05} & 2.73 \\
      & Adv.Test avg & 81.91 & \textbf{83.18} & 1.27 \\
      \bottomrule
    \end{tabular}
}
\caption{Performance gain when training larger models with our debiased datasets. Underscore indicates statistical significance against the baseline that is trained on the original datasets. For evaluation on SNLI-hard, the models are trained with SNLI or our debiased Seq-Z SNLI; for other evaluation datasets, the models are trained with MNLI or our debiased Z-Aug MNLI. Albert-xxlarge is experimented with one run due to its higher training cost.} \label{tab:stronger-models}
\end{center}
\end{table}

%================= Analysis =================
% \input{analysis}

%================= Related Works =================
% \pradeep{I'd recommend moving Related Work to the end of the paper, before Conclusion.}
% \ps{Sadly, I must agree. As much as I hate this fairly recent innovation in terms of flow.}
% \input{related}
\newacronym{poe}{PoE}{Product-of-Expert}

\section{Related Work}

\paragraph{Spurious Correlations in Datasets} The issue of spurious correlations in datasets between labels and simple input features has recently received significant attention~\citep{GururanganSLSBS18,PoliakNHRD18,belinkov-etal-2019-dont,karimi-mahabadi-etal-2020-end}. It has been shown that this issue is often inherent in the data annotation process, caused by biases in the framing of the task~\citep{schwartz-etal-2017-effect}, noisy annotations~\citep{chen-etal-2016-thorough}, or personal~\citep{geva-etal-2019-modeling} or group-level~\citep{liu2021toward} annotator biases. \citet{competency} provide a theoretical framework for analyzing spurious correlations, which we use to define our filtering mechanism in ~\cref{sec:z-filter}.

\paragraph{Debiasing NLI Models}
% Constructing unbiased datasets, especially at a large scale, is costly and challenging. Therefore, 
% 
Much prior work follows a \emph{model-centric} approach towards mitigating biases in NLI models -- they propose novel model architectures or training objectives to ensure that the models do not exploit the shortcuts presented by the dataset biases.
At the representation level, \citet{belinkov-etal-2019-dont,belinkov-etal-2019-adversarial} introduce an adversarial architecture to debias hypothesis representations to tackle hypothesis-only bias~\citep{GururanganSLSBS18}, and \citet{stacey-etal-2020-avoiding} strengthen the debiasing by using multiple adversarial classifiers.
\citet{zhou-bansal-2020-towards} use HEX projection to project the representation to the space orthogonal to the biased features to debias the model.
% \ps{Agreed, it is stated de facto, yet nothing is really ``said''.}
% model-level
At the model level, \citet{clark-etal-2019-dont,he-etal-2019-unlearn,karimi-mahabadi-etal-2020-end} propose methods based on \acrfull{poe}~\cite{poe} for mitigating biases by ensembling a biased-only model with a main model.
% and \citet{karimi-mahabadi-etal-2020-end} propose an end-to-end variant of \acrshort{poe}. 
\citet{utama-etal-2020-mind} propose the use of confidence regularization to improve out-of-distribution performance while retaining in-distribution accuracy. 
% weakness of model-centric approach
% These methods require changing the architecture or the learning objective of the model, and often lead to a more complex training procedure. \pradeep{This does not sound like a strong argument against model-centric approaches. Can we say something about the changes not generalizing across models or model-classes?}
% \ps{How about something akin to: ``\ldots given this, it is necessary to adapt it between model categories and thus additional effort on the part of model designers? Plus, hyper parameter tuning!'' Still, is it really necessary to delve on this that much? If we have better results, we have better results, no need to argue much about it. You know what they say, talking shit about others kind of makes you look bad as well. ;P How about we just list things as advantages after we have talked about our own approach? That feels more ``positive''.}

\paragraph{Debiasing NLI Datasets}
% Prior work towards debiasing NLI datasets \emph{perturb} individual instances as a data augmentation strategy. 
\citet{ross2021tailor} introduce TAILOR, a semantically-controlled perturbation method for data augmentation based on a small number of manually defined perturbation strategies.
\citet{AFLite} propose AFLite, a dataset filtering method that learns feature representations with a model and conduct adversarial filtering based on model predictions. 
Unlike these approaches, our method requires no manually-written perturbation heuristics and is model-agnostic, hence it is more generally applicable.

\paragraph{Generative Data Augmentation}
Several works investigate generative data augmentation techniques to improve model robustness in other areas.
\citet{G-DAug} conduct generative data augmentation for commonsense reasoning and show that it can improve out-of-domain generalisation.
\citet{lee2021crossaug} trains a generator to generate new claims and evidence for debiasing fact verification datasets like FEVER~\citep{fever}.
\citet{DINO} exploit large pretrained language models to generate semantic textual similarity datasets.
\citet{bartolo2021improving} improve robustness of question answering models by generating adversarial dataset.

\section{Conclusions}
To address the issue of spurious correlations between task-independent features and labels in NLI datasets, we propose methods to generate label-consistent data and then filter out instances from existing datasets that contribute to those spurious correlations; thereby generating debiased datasets.
Models trained on our debiased versions of the SNLI and MNLI datasets generalise better than the equivalent model trained on the original datasets to a large suite of test sets focusing on various kinds of known biases.
%
% The analysis also shows that \yw{[TODO]}.
%
Future work in this direction includes investigating whether our techniques are applicable to tasks beyond NLI.
%
% \ps{
%     Do we really need some sort of token future work sentence? Does anyone care these days?
% }

\section*{Acknowledgments}

The authors would like to thank Max Bartolo, Alexis Ross, Doug Downey, Jesse Dodge, Pasquale Minervini, and Sebastian Riedel for their helpful discussion and feedback.

\bibliographystyle{acl_natbib}
\bibliography{main}

\clearpage

\appendix

\section{Hyperparameters}\label{appendix:hparams}

\subsection{Hyperparameters of Our Proposed Method}

\begin{table}[!hb]
\begin{center}
% \resizebox{\columnwidth}{!}{
    \begin{tabular}{lc}
      \toprule
      {\bf Hyperparameter} & {\bf Value} \\
      \midrule
      learning rate & 1e-5 \\
      batch size & 24 \\
      epoch & 5 \\
      optimiser & Adam \\
      Adam $\epsilon$ & 1e-6 \\
      Adam $(\beta_1, \beta_2)$ & (0.9, 0.999) \\
      learning rate scheduler & constant \\
      max sequence length & 128 \\
      pretrained model & \href{https://huggingface.co/gpt2-large}{GPT-2 large} \\
      device & Nvidia A100 \\
      $\lambda$ & 0.5 \\
      $\alpha$ & 1.0 \\
      \bottomrule
    \end{tabular}
% }
\caption{Hyperparameters for training the generator $G^*$.}
\end{center}
\end{table}

\begin{table}[!hb]
\begin{center}
\resizebox{\columnwidth}{!}{
    \begin{tabular}{lc}
      \toprule
      {\bf Hyperparameter} & {\bf Value} \\
      \midrule
      number of samples from $G^*_{\text{SNLI}}$ & 5,000,000 \\
      number of samples from $G^*_{\text{MNLI}}$ & 4,000,000 \\
      data filtering threshold $\tau$ & 0.95 \\
      data filtering model &  \href{https://huggingface.co/roberta-large}{Roberta-large}  \\
      z-filtering number of biased features & 20 \\
      \bottomrule
    \end{tabular}
}
\caption{Hyperparameters of the data generation pipeline.}
\end{center}
\end{table}

\begin{table}[!hb]
\begin{center}
\resizebox{\columnwidth}{!}{
    \begin{tabular}{lc}
      \toprule
      {\bf Hyperparameter} & {\bf Value} \\
      \midrule
      learning rate & 1e-5 \\
      batch size & 32 \\
      epoch & 5 \\
      optimiser & Adam \\
      Adam $\epsilon$ & 1e-6 \\
      Adam $(\beta_1, \beta_2)$ & (0.9, 0.999) \\
      learning rate scheduler & constant with warmup \\
      warm up steps & 2000 \\
      max sequence length & 128 \\
      pretrained model & \href{https://huggingface.co/bert-base-uncased}{BERT-base} \\
      device & Nvidia A100 \\
      early stop patience & 3 epochs \\
      \bottomrule
    \end{tabular}
}
\caption{Hyperparameters for training the NLI models.}
\end{center}
\end{table}

\subsection{Hyperparameter Tuning of PoE}
The learning objective of PoE is defined as follows:
\begin{equation*}
    \mathcal{L}_{\text{PoE}} =  \sum_{i=1}^{|\mathcal{D}|} CE(l_i, p'_i) + \gamma CE(l_i, b_i),
\end{equation*}
where $CE$ stands for cross-entropy loss, $l_i$ is the label, and $\gamma$ is a hyperparameter. $p'_i=softmax(\log p_i + \beta \log b_i)$ is the ensemble of the main model's prediction $p_i$, and the bias-only model's prediction $b_i$ weighted by a hyperparameter $\beta$.

We find that the result of PoE is very sensitive to the hyperparameters $\beta$ and $\gamma$. Following~\citet{karimi-mahabadi-etal-2020-end}, we conduct grid search for the two hyperparameters, with $\beta \in \{0.05,0.1,0.2,0.4,0.8,1.0,2.0\}$ and $\gamma \in \{0.05,0.1,0.2,0.4,0.8,1.0\}$. The best hyperparameters found for each evaluation dataset is listed in~\cref{tab:poe-hparams}.

\begin{table}[!hbt]
\begin{center}
% \resizebox{\columnwidth}{!}{
    \begin{tabular}{lccc}
      \toprule
      {\bf Train data} & {\bf Eval. data} & {\bf $\beta$} & {\bf $\gamma$}  \\
      \midrule
      SNLI & SNLI-hard & 2.0 & 0.4 \\
      \midrule
      Seq-Z SNLI & SNLI-hard & 2.0 & 0.4 \\
      \midrule
    %   PoE w/ MNLI & MNLI-m hard & 0.8 & 1.0 \\
    %   PoE w/ Z-Aug MNLI  & MNLI-m hard & 2.0 & 0.4 \\ \hline
    %   PoE w/ MNLI  & MNLI-mm hard & 2.0 & 0.4 \\
    %   PoE w/ Z-Aug MNLI  & MNLI-mm hard & 2.0 & 0.8 \\ \hline
    %   PoE w/ MNLI & HANS & 2.0 & 0.8 \\ 
    %   PoE w/ Z-Aug MNLI  & HANS & 2.0 & 1.0 \\ \hline
      
      MNLI & 
        \begin{tabular}{@{}c@{}c@{}} MNLI-m hard \\MNLI-mm hard \\ HANS\end{tabular} & 
        \begin{tabular}{@{}c@{}c@{}} 0.8 \\ 2.0 \\ 2.0 \end{tabular} & 
        \begin{tabular}{@{}c@{}c@{}} 1.0 \\ 0.4 \\ 0.8 \end{tabular}\\
      \midrule
      Z-Aug MNLI  & 
        \begin{tabular}{@{}c@{}c@{}} MNLI-m hard \\MNLI-mm hard \\ HANS\end{tabular} & 
        \begin{tabular}{@{}c@{}c@{}} 2.0 \\ 2.0 \\ 2.0 \end{tabular} & 
        \begin{tabular}{@{}c@{}c@{}} 0.4 \\ 0.8 \\ 1.0 \end{tabular}\\
      \bottomrule
    \end{tabular}
% }
\caption{Best hyperparameters found for PoE models with different training and evaluation datasets.} \label{tab:poe-hparams}
\end{center}
\end{table}

\section{Task-independent Features} \label{appendix:bias-features}

We list the chosen set of task-independent features that we aim to mitigate in this work in \cref{tab:feature-desc}.
Note that our method does not depend on the choice of task-independent features.
One can easily add their own features in the future to mitigate newly-identified spurious correlations.

\cref{tab:biased-features} shows the most salient task-independent features (ranked by z-statistics) in SNLI and our debiased SNLI dataset. It shows that the correlation between task-independent features and labels is massively reduced, dropping from over 400 to roughly 17. These results verify that our method successfully mitigates the spurious correlations in the dataset.

\begin{table*}[htb]
\begin{center}
\resizebox{\textwidth}{!}{
    \begin{tabular}{ll}
      \toprule
      
      {\bf Feature} & {\bf Description} \\ \midrule \midrule
      
      {Unigrams \& Bigrams} & All unigrams and bigrams. The n-grams from premise and hypothesis are treated separately. \\ \midrule
      
      {Hypothesis length} & Number of tokens in the hypothesis. \\ \midrule
      
      {Hypothesis-premise length ratio} & {Number of tokens in hypothesis divided by number of tokens in the premise.} \\ \midrule
      
      {Lexical overlap} & Ratio of tokens in the hypothesis that overlap with the premise. \\ \midrule
      
      {Hypothesis-only model's prediction} & We train a hypothesis-only model on the original dataset and use its prediction as a feature. \\ \midrule
      
      {Null feature} & A dummy feature added for \emph{all} instances to avoid skewed label distribution. \\
      
      \bottomrule
    \end{tabular}
}
\caption{Descriptions of the features used to debias the datasets in \cref{sec:fixing_distribution}.} \label{tab:feature-desc}
\end{center}
\end{table*}

\begin{table}[hbt]
\begin{center}
\resizebox{\columnwidth}{!}{
    \begin{tabular}{lclc}
      \toprule
      \multicolumn{2}{c}{\bf{SNLI}} & \multicolumn{2}{c}{\bf{Debiased SNLI (Seq-Z)}} \\
      {\bf Biased feature} &  {\bf z-statistics}  & {\bf Biased feature} &  {\bf z-statistics}  \\
      \midrule
      \multicolumn{4}{l}{\bf{Entailment}} \\
        hypo-only-pred=0 & 422.1  & theres@hypothesis & 17.5  \\
        lex-overlap$>0.8$ & 123.3  & hypo-len$<5$ & 17.4 \\
        full-lex-overlap & 117.3 & full-lex-overlap & 17.4 \\
        outside@hypothesis & 102.2 & politician@hypothesis & 17.4 \\
        lex-overlap$>0.9$ & 90.4 & speaking@hypothesis & 17.4 \\
      \midrule
      \multicolumn{4}{l}{\bf{Neutral}} \\
    hypo-only-pred=1 & 436.1  &  championship@hypothesis & 15.3 \\
    for a@hypothesis & 63.6 &  living room@hypothesis & 15.2 \\
    his@hypothesis & 56.8 & many men@hypothesis & 15.2 \\
    friends@hypothesis & 55.6  & green suit@hypothesis & 15.2 \\
    tall@hypothesis & 52.7 & are wearing@hypothesis & 15.2 \\
      \midrule
      \multicolumn{4}{l}{\bf{Contradiction}} \\
    hypo-only-pred=2 & 433.9 & nothing@hypothesis & 17.0 \\
    sleeping@hypothesis & 92.9 & hypo-only-pred=2 & 16.9 \\
    is sleeping@hypothesis & 68.7 & at home@hypothesis & 16.9 \\
    nobody@hypothesis & 68.4 & is no@hypothesis & 16.9 \\
    no@hypothesis & 62.7 & york yankees@hypothesis & 16.9 \\
      \bottomrule
    \end{tabular}
}
\caption{Top-5 biased features with the highest z-statistics on SNLI (left) and debiased Seq-Z SNLI (right) for each label class.} \label{tab:biased-features}
\end{center}
\end{table}

\section{Description of Adversarial Test~\citep{model-agnostic-debias} Subcategories} \label{appendix:adv-test-desc}

% We list the set of bias features that we mitigate in this work in \cref{tab:adv-test-desc}.

The adversarial test benchmark~\citep{model-agnostic-debias} includes the following subcategories from various sources:
\begin{itemize}
    \item PI-CD: classifier detected partial-input~\citep{GururanganSLSBS18}.
    \item PI-SP: HypoNLI~\citep{liu-etal-2020-hyponli} dataset that tackles surface patterns heuristics.
    \item IS-SD: syntactic diagnostic dataset HANS~\citep{hans}.
    \item IS-CS: lexically misleading instances constructed by~\citet{nie2019analyzing}.
    \item LI-LI: lexical inference test by \citep{naik-etal-2018-stress,glockner-etal-2018-breaking}.
    \item LI-TS: text-fragment swap test by swapping the premise and hypothesis~\citep{wang2019if,minervini-riedel-2018-adversarially}.
    \item ST: an aggregation of word-overlap (ST-WO), negation (ST-NE), length mismatch (ST-LM), and spelling errors (ST-SE) tests in~\citep{naik-etal-2018-stress}.
\end{itemize}

\section{Visualisation of z-statistics}

\begin{figure*}[t]
\begin{center}
\begin{subfigure}{.49\textwidth}
\includegraphics[width=\textwidth]{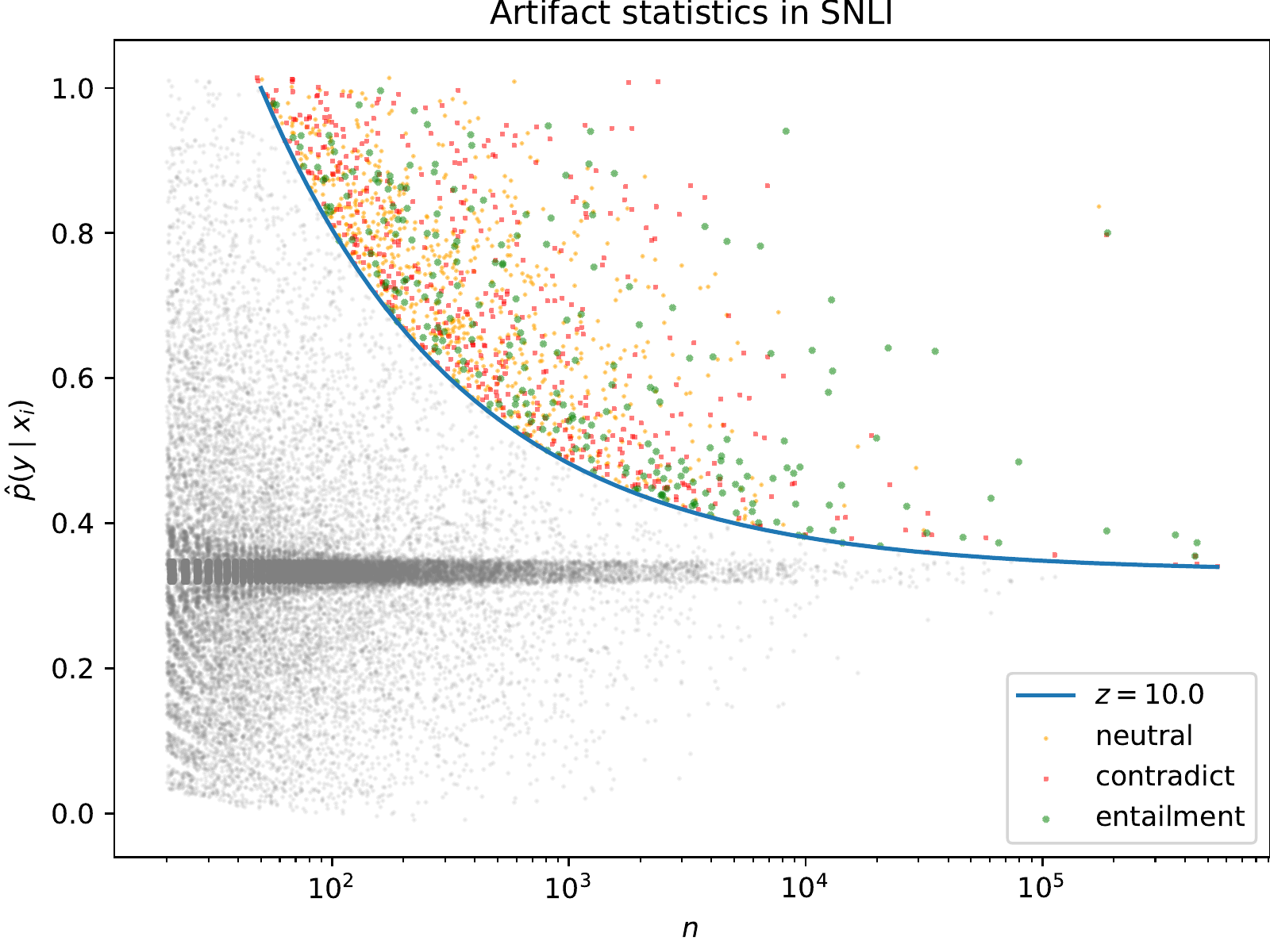}
% \subcaption{\globalmethodRL{} vs. baselines} \label{fig:master}
\end{subfigure}
\begin{subfigure}{.49\textwidth}
\includegraphics[width=\textwidth]{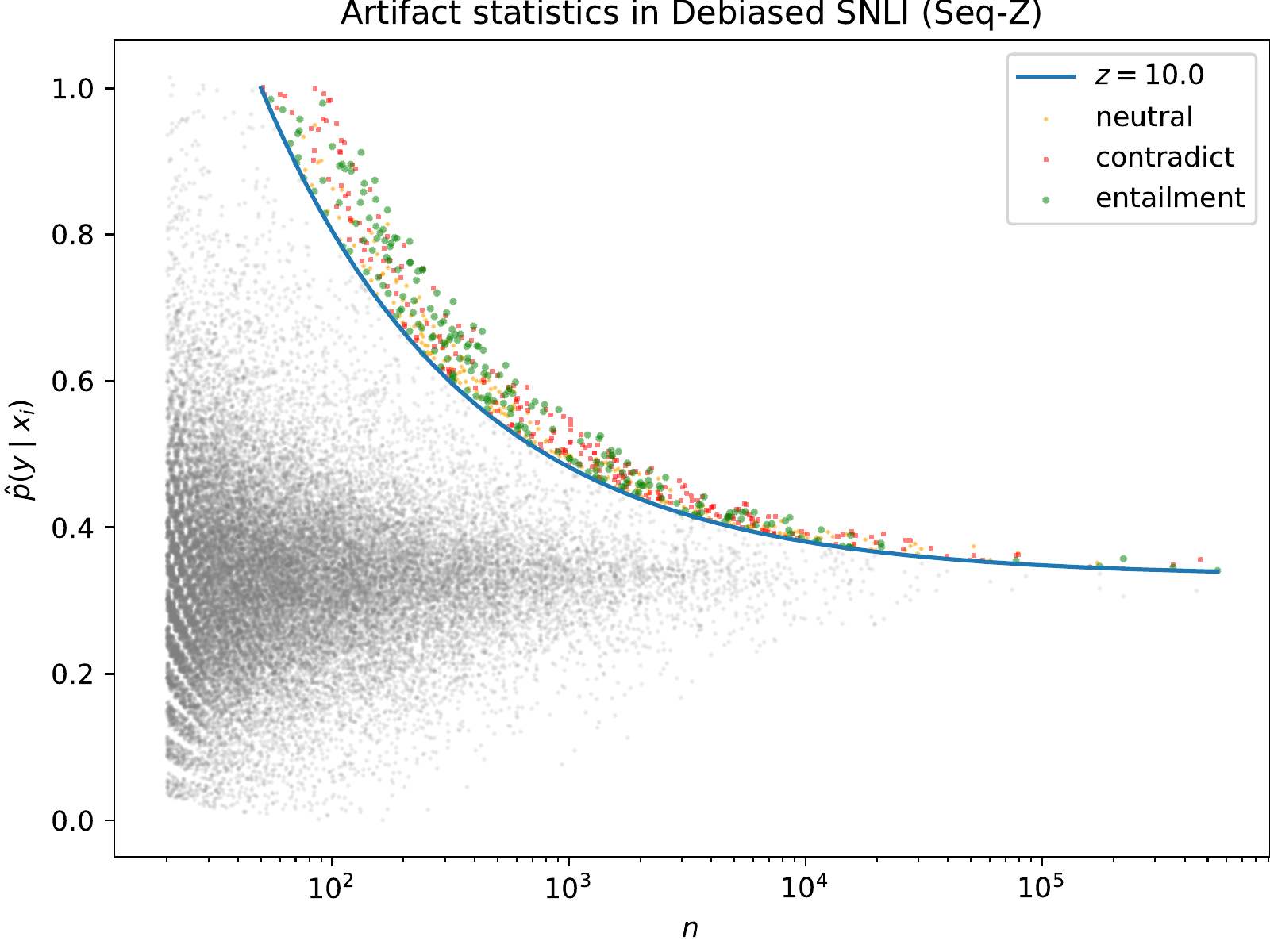}
% \subcaption{Local vs. Global Models} \label{fig:local_vs_global}
\end{subfigure}
\end{center}
\caption{Statistics of the features on SNLI and our debiased SNLI (Seq-Z).} \label{fig:z-plot}
\end{figure*}

Following~\citet{competency}, we visualise the statistics of the features on both SNLI and our debiased SNLI (Seq-Z) dataset in~\cref{fig:z-plot}.\footnote{We sample 10\% of the points under the $z=10.0$ curve to compress the figure, but it may still be slow to render the figures because the number of points is still large.}
Comparing the two plots, it confirms that our method successfully suppresses the spurious correlations in the dataset.

\section{Ablation Study} \label{sec:ablation}

\begin{table}[hbt]
\begin{center}
\resizebox{\columnwidth}{!}{
    \begin{tabular}{lccc}
      \toprule
      Data & Size & SNLI & SNLI-hard  \\
      \midrule
      Seq-Z $\mathcal{Z}( \hat{\mathcal{D}}_{G^*} | \mathcal{Z}(\mathcal{D}_{\text{SNLI}}))$ & 928k & 88.08 & 82.82$_{\pm 0.15}$ \\
      Seq-Z $\mathcal{Z}( \hat{\mathcal{D}}_{G^*} | \mathcal{Z}(\mathcal{D}_{\text{SNLI}}))$ & 549k & 87.59 & 82.35$_{\pm 0.46}$ \\
      Seq-Z $\mathcal{Z}( \hat{\mathcal{D}}_{G} | \mathcal{Z}(\mathcal{D}_{\text{SNLI}}))$ & 549k & 88.15 & 82.20$_{\pm 0.23}$ \\
      \hline
      $\mathcal{D}_{\text{SNLI}} \cup \hat{\mathcal{D}}_{G^*} $ & 2577k & 90.85 & 81.99$_{\pm 0.47}$ \\
      $\mathcal{D}_{\text{SNLI}} \cup \hat{\mathcal{D}}_{G} $ & 3717k & 90.83 & 80.82$_{\pm 0.27}$ \\
      \hline
      Z-Aug $\mathcal{Z}(\hat{\mathcal{D}}_{G^*} | \mathcal{D}_{\text{SNLI}})$ & 1142k & 90.67 & 81.78$_{\pm 0.53}$ \\
      $\mathcal{D}_{\text{SNLI}} \cup \hat{\mathcal{D}}_{G^*} $ & 1142k & 90.72 & 81.45$_{\pm 0.52}$ \\
      $\mathcal{D}_{\text{SNLI}} \cup \hat{\mathcal{D}}_{G} $ & 1142k & 90.67 & 80.85$_{\pm 0.27}$ \\
      \hline
      $\mathcal{Z}(\hat{\mathcal{D}}_{G^*})$ & 808k & 88.44 & 81.28$_{\pm 0.57}$ \\
      $\mathcal{Z}(\hat{\mathcal{D}}_{G^*})$ & 549k &  88.12 & 80.67$_{\pm 0.41}$ \\
      $\hat{\mathcal{D}}_{G^*}$ (w/ filter) & 549k &  88.59 & 80.41$_{\pm 0.50}$  \\
      $\mathcal{D}_{G^*}$ (wo/ filter) & 808k & 75.65 & 76.67$_{\pm 0.83}$ \\
      $\mathcal{D}_{G^*}$ (wo/ filter) & 549k & 75.43 & 76.05$_{\pm 0.49}$ \\
      $\mathcal{Z}(\mathcal{D}_{\text{SNLI}})$ & 127k & 84.93 & 80.52$_{\pm 1.03}$ \\
      original SNLI $\mathcal{D}_{\text{SNLI}}$ & 549k  & 90.45 & 80.34$_{\pm 0.46}$ \\
      \bottomrule
    \end{tabular}
}
\caption{Ablation study conducted on SNLI and SNLI-hard.} \label{tab:ablation-snli}
\end{center}
\end{table}

\section{Generated Samples of Debiased Dataset}

\cref{tab:samples-snli} and~\cref{tab:samples-mnli} show generated samples in the debiased SNLI and MNLI datasets respectively.
The samples are quite diverse and the quality is reasonably good, which demonstrates the effectiveness of our quality ensuring techniques presented in~\cref{sec:data-quality}.

\begin{table*}[htb]
\begin{center}
\resizebox{\textwidth}{!}{
    \begin{tabular}{p{12cm}p{7.5cm}l}
      \toprule
      
      {\bf Premise} & {\bf Hypothesis} & {\bf Label}\\ \midrule \midrule
      
Thanksgiving dinner is a fun time for everyone. & The dinner is a fun event. & entailment \\ \hline
A father is letting his toddler drink from his glass. & A toddler is having a drink & entailment \\ \hline
Hair stylist performing a haircut. & A hailer is performing surgery & contradiction \\ \hline
Then there are two men in white shirts, one of which is holding a cigarette and the other an open book. & Two men sit at a conference table with a book and a cigarette. & neutral \\ \hline
Three men playing basketball on a court with an audience in the background. & Three people playing basketball & entailment \\ \hline
Six children, boys and girls, jumping into a swimming pool. & Six children are jumping into a pool & entailment \\ \hline
Three girls jump for joy in front of a building. & The kids are sitting on their front steps. & contradiction \\ \hline
The child in the green one piece suit is running in the playground. & The child is playing outside & entailment \\ \hline
View of an intersection with city buses and a police car. & The intersection is surrounded by vehicles. & entailment \\ \hline
The man on the yellow basketball team tries to score while the men on the opposing team try to block his shot. & Two men on different teams are competing in a game of a male is trying to score while other men on the opposing one defend his basket in basketball & entailment \\ \hline
Young child wearing orange shirt eating a ice cream cone. & A child eats ice cream at the ice cream stand. & neutral \\ \hline
Five people standing in front of a shopping center. & Five people outside the building & entailment \\ \hline
He's taking a break after a long workout. & He is taking a break from his workout & entailment \\ \hline
Two men are sitting on a couch, playing music together. & The two people play guitars. & neutral \\ \hline
Everyone is out enjoying the winter weather and having fun with their children. & Everyone is out enjoying the summer & contradiction \\ \hline
Many people walking through a city street. & There are a group of people in Times Square. & neutral \\ \hline
A woman in a black dress walks down the street. & a person in dresses walks & entailment \\ \hline
Four children, riding unicycles, are on a sidewalk in front of a brick building. & Four children ride unicycles on the sidewalk & entailment \\ \hline
Four kids playing soccer in a field. & The children played with bubbles. & contradiction \\ \hline
MADISON, Wis. (AP) — The man in the white jersey and orange visor threw the ball for the two boys in uniforms with blue jerseys. & A man in white is throwing a ball to two boys in blue uniforms. & entailment \\ \hline
Mikhail Kasyapkin, who plays Bart on The Simpsons, is talking to a woman. & The woman tells him to stop making couples sit & neutral \\ \hline
Shutterstock photo of a woman with a heart tattoo on her calf. & A woman with a pumpkin tattoo on her back & contradiction \\ \hline
Three women and a man sing their hearts out in the microphone. & A group singing & entailment \\ \hline
With so many people on the beach, the woman in yellow has to make a quick decision. & Many people are at a beach, one has to make a decision & entailment \\ \hline
Celebrants are walking with American flags. & People are walking. & entailment \\ \hline
Customer examines flowers at a market. & A customer examines flowers. & entailment \\ \hline
He is in the air on his skateboard. & A guy is in a tree. & contradiction \\ \hline
thousands of people enjoying a fireworks show. & There is an audience for a show. & entailment \\ \hline
Bicyclists in a race, with a blue bike leaving the ground in the lead. & Bikers resting after a long ride. & contradiction \\ \hline
He has a pet bird in a cage, and it is sleeping. & He is walking the dogs. & contradiction \\ \hline

      \bottomrule
    \end{tabular}
}
\caption{Generated samples in the debiased SNLI datasets.} \label{tab:samples-snli}
\end{center}
\end{table*}

\begin{table*}[htb]
\begin{center}
\resizebox{\textwidth}{!}{
    \begin{tabular}{p{11cm}p{9cm}l}
      \toprule
      
      {\bf Premise} & {\bf Hypothesis} & {\bf Label}\\ \midrule \midrule
      As I noted earlier, the board and the auditors should have a strategic alignment of interests. & The board should align to increase efficiency. & neutral \\ \hline
This story was originally published in Slate. For more on the U.S. role in that war, subscribe to Slate's Subscribe now! & The U.S. played very little part in the war. & neutral \\ \hline
Via Newsday's  a poll finds that 84 percent of Americans think Monica Lewinsky should tell the truth about her encounter with Clinton. & A majority of the public thinks Lewinsky should come forward. & entailment \\ \hline
Violence among theatrical people, on the other hand, can be entertainingly savage, cf, All About Eve (1884) and The Mousetrap (1928). & There aren, always hasn't usually been oancy situation with violence among theatrical people because they don't have to work because it isn't employment. & contradiction \\ \hline
Nowhere in the book does Hatfield warn the reader that he has altered details or created composite characters to protect his sources. & Hatfield didn't inform the readers in any part in the book that the details of the altered information was to protect his sources & entailment \\ \hline
The young inhabitants are brought up knowing nothing else. & The young inhabitants have been brought up knowing of nothing. & entailment \\ \hline
The 5th floor of the Royal Palace is open to the public, with restricted access for foreign guests. & Foreign guest have restricted access in the royal palace for visitors. & entailment \\ \hline
Pulitzer Prizes are given to books, magazines, paintings, and sculpture. & You won a prize when you eat blueberries at dinner. & contradiction \\ \hline
I admit I didn't have much reason to think that. & After all, most of the people don;t think that way. & neutral \\ \hline
In the past, Medicare's fiscal health has generally been gauged by the solvency of the HI trust fund projected over a 75-year period. & Medicare's soliesic fitness is displayed in the form of the surplus projected over a 50 year term. & contradiction \\ \hline
% No, wait, that's what we knew yesterday about England, although it's not far off. & We thought that England had it the worst of everything. & neutral \\ \hline
A case study where the only people interviewed were senior officials would be seen as a not-good case study, in contrast to one where the views of individuals at all levels affected was obtained. & If senior editors were interviewed they would not be considered the best examples for case studies. & entailment \\ \hline
If you've ever spent an evening plunging your wrists into ice water, you are an easy mark for devices that promise to relieve carpal tunnel syndrome. & People are easy marks for devices that may cure cat paral tunnel syndrome & entailment \\ \hline
It's a sign of a permanently altered world that natural blondness should have such sacred power no longer. & The people still believe blondness has a special significance. & contradiction \\ \hline
% It was Number Six of my catalogue." & It was sixth in my catalog & entailment \\ \hline
The Three-Arched Bridge, by Ismail Kadare, translated by John Hodgson (Arcade). & Ismail Marare translated The Three-Aral. & contradiction \\ \hline
% Click to read the best of the nominations. & Click to view the author's comments on the writing's popularity. & contradiction \\ \hline
Many of these organizations found themselves in an environment similar to the one confronting federal managers today-one in which they were called upon to improve performance while simultaneously reducing costs. & This was the only option for all their group. & neutral \\ \hline
The long-sought, the elusive, the elusive Jane Finn! & She is easily obtainable. & contradiction \\ \hline
And now, to-day, he puts forward a suggestion that he himself must have known was ridiculous. & He is making the ridiculous suggestion that himself must have been aware of. & entailment \\ \hline
Jupiter's moon, Callisto, has a thick atmosphere and is a good destination for a quiet tour. & Callisto's atmosphere makes for a pleasant journey to explore. & entailment \\ \hline
Founded in 1995, the Agora formed to address the enormous security challenges brought about by new computer, network, and Internet technologies. & The Agora was formed to address the challenge of nuclear proliferation. & contradiction \\ \hline
Just last week in The New Yorker, Malcolm Gladwell argued that Gen. & Just last week in Newsweek, Johnny Chung argued that Gen. & contradiction \\ \hline
Muller and most of the boys can be counted on not to cause any more than the normal pay-night disturbances. & Muller will not start a fist fight. & neutral \\ \hline
Don't call me Shirley. & My last name is Shirley and that is how I want to be referred to. & contradiction \\ \hline
The vast majority of the approximately 1,700 lawyers at LSC-funded programs around the country volunteer for only a single case, whether it is a class action suit, a simple civil rights case or a case involving a dangerous person. & There's no reason to get one or do the work otherwise. & neutral \\ \hline
The Promise Keepers talk far less about abortion and homosexuality than their critics and the media do. & They're surrounded far less with the issues that the media and other critics deal with. & entailment \\ \hline
It was Susan in his head. & Susan was telling him exactly to his surprise. & neutral \\ \hline
% It was already there ”in the mixture. & There was nothing to look for in order to be in the chemists assistant's shop when we arrived. & neutral \\ \hline
In 1782, after only a few years, the city decided to impose planning guidelines. & It took a few decades for 17 year-olds. & contradiction \\ \hline
      \bottomrule
    \end{tabular}
}
\caption{Generated samples in the debiased MNLI datasets.} \label{tab:samples-mnli}
\end{center}
\end{table*}

\end{document}